\pdfoutput=1

\documentclass[11pt]{article}

\usepackage{acl}
\usepackage{times}
\usepackage{latexsym}

\usepackage[T1]{fontenc}

\usepackage[utf8]{inputenc}

\usepackage{microtype}

\usepackage{algorithm}
\usepackage{algorithmic}
\usepackage{array}
\usepackage{mathtools}
\usepackage{amsmath}
\usepackage{dsfont}
\usepackage{subfigure}
\usepackage{inconsolata}
\usepackage{graphicx}
\usepackage{multicol}
\usepackage{multirow}
\usepackage{xcolor}
\usepackage{colortbl}
\definecolor{Gray}{gray}{0.93}
\usepackage{amssymb}
\usepackage{amsthm}
\usepackage{enumitem}
\usepackage{bm}
\usepackage{booktabs}
\usepackage{units}
\usepackage{bbm}
\usepackage{tabu}
\usepackage{fontawesome}
\usepackage{pifont}
\usepackage{tcolorbox}
\usepackage{colortbl}
\usepackage{xcolor}
\newcommand{\ourmethod}{Qwen2.5-Coder-Instruct-C}
\newcommand{\benchmark}{\textsc{ExecRepoBench}}
\newcommand{\instruct}{\textsc{Repo-Instruct}}
\newcommand\identity{1\kern-0.25em\text{l}}

\title{\benchmark{}: Multi-level Executable Code Completion Evaluation}

\author{
  Jian Yang\textsuperscript{\rm 1},  
  {\bf Jiajun Zhang}\textsuperscript{\rm 2}, 
  Jiaxi Yang\textsuperscript{\rm 3}, 
  {\bf Ke Jin}, 
  {\bf Lei Zhang}\textsuperscript{\rm 3},
  {\bf Qiyao Peng}, \\
  {\bf Ken Deng}\textsuperscript{\rm 1},
  {\bf Yibo Miao}\textsuperscript{\rm 4}, 
  {\bf Tianyu Liu}\textsuperscript{\rm 1}, 
  {\bf Zeyu Cui}\textsuperscript{\rm 1}, 
  {\bf Binyuan Hui}\textsuperscript{\rm 1},
  {\bf Junyang Lin}\textsuperscript{\rm 1}, \\
  \textsuperscript{\rm 1}Alibaba Group; 
  \textsuperscript{\rm 2}University of Science and Technology of China; \\
  \textsuperscript{\rm 3}University of Chinese Academy of Sciences;
  \textsuperscript{\rm 4}Shanghai Jiao Tong University \\
  \{yj411294,binyuan.hby,junyang.ljy\}@alibaba-inc.com; \\ 
}

\begin{document}
\maketitle
\begin{abstract}
Code completion has become an essential tool for daily software development. Existing evaluation benchmarks often employ static methods that do not fully capture the dynamic nature of real-world coding environments and face significant challenges, including limited context length, reliance on superficial evaluation metrics, and potential overfitting to training datasets. In this work, we introduce a novel framework for enhancing code completion in software development through the creation of a repository-level benchmark \benchmark{} and the instruction corpora \instruct{}, aim at improving the functionality of open-source large language models (LLMs) in real-world coding scenarios that involve complex interdependencies across multiple files. \benchmark{} include 1.2K samples from active Python repositories. Plus, we present a multi-level grammar-based completion methodology conditioned on the abstract syntax tree to mask code fragments at various logical units (e.g. statements, expressions, and functions). Then, we fine-tune the open-source LLM with 7B parameters on \instruct{} to produce a strong code completion baseline model \ourmethod{} based on the open-source model. \ourmethod{} is rigorously evaluated against benchmarks, including MultiPL-E and \benchmark{}, which consistently outperforms prior baselines across all programming languages. The deployment of \ourmethod{} can be used as a high-performance, local service for programming development\footnote{\url{https://execrepobench.github.io/}}.
\end{abstract}

\section{Introduction}
In the field of software engineering, the emergence of large language models (LLMs) designed specifically for code-related tasks has represented a significant advancement. These code LLMs~\cite{AlphaCode,santacoder}, such as DeepSeek-Coder~\cite{deepseek_coder} and Qwen-Coder~\cite{qwen25coder}, have been pre-trained on extensive datasets comprising billions of code-related data. The advent of code LLMs has revolutionized the automation of software development tasks, providing contextually relevant code suggestions and facilitating code generation.

\begin{figure}[t]
\centering
\includegraphics[width=1.0\linewidth]{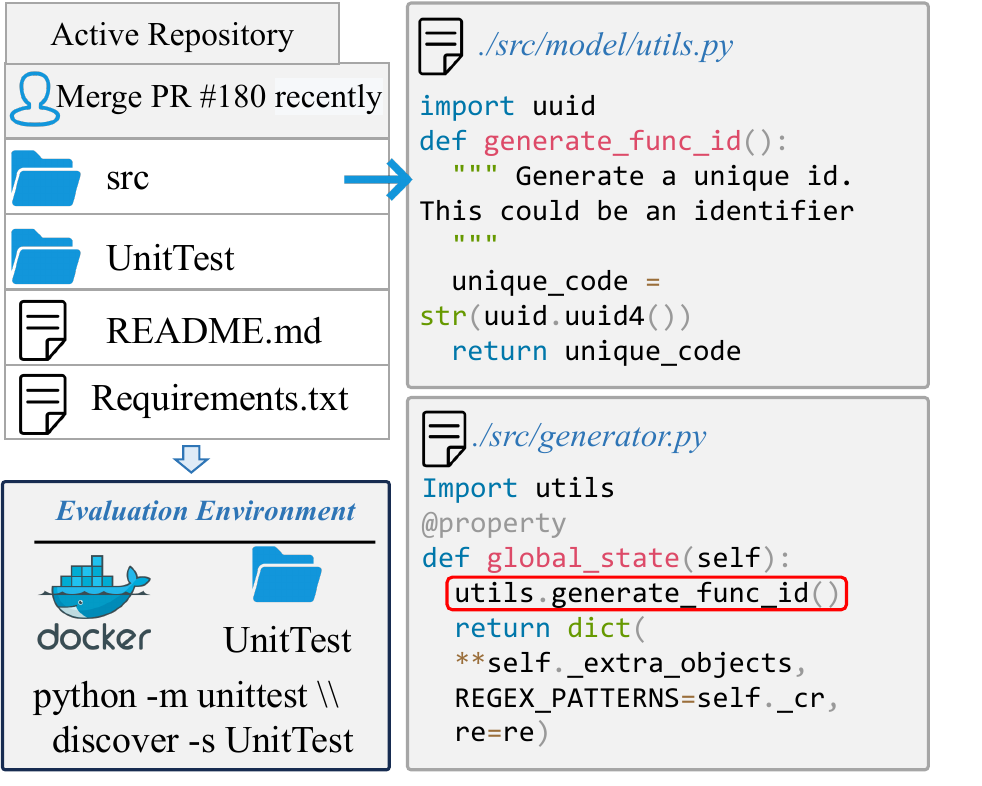}
\caption{Executable Repository-level code evaluation with the given test cases.}
\vspace{-10pt}
\label{intro}
\end{figure}

The code completion task holds paramount importance in modern software development, acting as a cornerstone for enhancing coding efficiency and accuracy. By analyzing the context of the ongoing work and using sophisticated algorithms to predict and suggest the next segments of code, code completion tools drastically reduce the time and effort programmers spend on writing boilerplate code, navigating large codebases, or recalling complex APIs and frameworks, which both accelerates the software development cycle and significantly diminishes the likelihood of syntax errors and bugs, leading to cleaner, more maintainable code. The recent code LLMs \citep{fim,codegeex} complete the middle code based on the prefix and suffix code through prefix-suffix-middle (PSM) and suffix-prefix-middle (SPM) pre-training paradigm. To correctly evaluate the code completion capability, the HumanEval benchmark \citep{santacoder,codegeex} is extended to the infilling task by randomly masking some code spans and lines and prompting LLms to predict the middle code. The recent works \citep{crosscodeeval,cocomic,ding2023static} propose to use the cross-file context to complete the current file and then score the results with $n$-gram string match. \textit{However, the community still lacks an executable evaluation repository-level benchmark from live repositories and the corresponding instruction corpora.}

In this work, we benchmark, elicit, and enhance code repository-level completion tasks of open-source large language models (LLMs) by creating the repository-level instruction corpora \textbf{\instruct{}} and the corresponding benchmark \textbf{\benchmark{}} for utilization and evaluation for code completion in real-world software development scenarios, where projects frequently involve complex dependencies across multiple files. Unlike previous benchmarks with text-matching metrics (e.g. exact match (EM) and edit similarity (ES)), we \benchmark{} is constructed with repository-level unit tests to verify the correctness of the completion code, which contains 1.2K samples from 50 active Python repositories. To facilitate the attention of the community for the code completion task, we propose the multi-level grammar-based completion to create \instruct{}, where the code fragments under the different levels of logical units are masked for completion using the parsed abstract syntax tree (AST). During supervised finetuning (SFT), the code snippet of the repository is packed into the instruction data for the code completion LLMs \ourmethod{}, where the query gives the prefix code of the current file, suffix code of the current file, and code snippets of other files.

\ourmethod{} is evaluated on the code generation benchmark \cite{multipl_e} and our created code completion benchmark \benchmark{}. The results demonstrate that \ourmethod{} consistently achieves state-of-the-art performance across all languages, notably surpassing the previous baselines.
The contributions are summarized as follows:
\begin{itemize}
\setlength\itemsep{0em}
    \item We introduce executable repository-level benchmark \benchmark{} for code completion evaluation, which collects the active repositories from GitHub and modify them into 
    executable formats with test cases.
    \item We propose the multi-level grammar-based completion conditioned on the abstract syntax tree, where the statement-level, expression-level, function-level, and class-level code snippets are extracted for multi-level completion instruction corpora \instruct{}
    \item Based on the open-source LLMs and the instruction corpora \instruct{}, we fine-tune base LLMs with 7B parameters \ourmethod{} with a mixture of code completion data and standard instruction corpora, which can be used as a local service for programming developer.
\end{itemize}

\begin{figure}[t]
\begin{center}
\includegraphics[width=1.0\columnwidth]{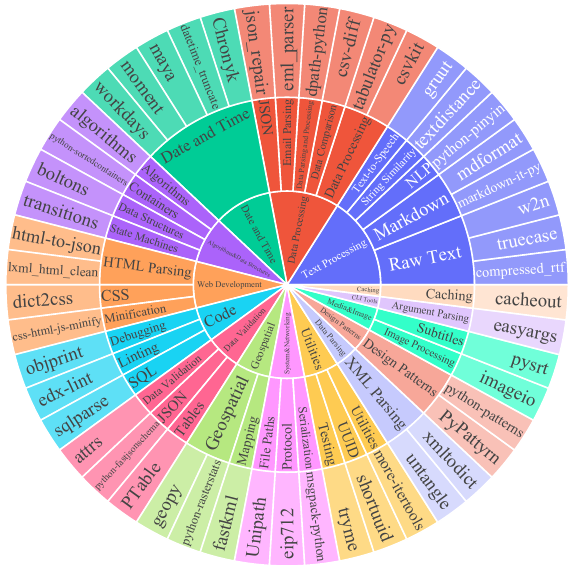}
	\caption{Classifiction of collected repositories.}
	\label{repo_classification}
\end{center}
\end{figure}

\begin{figure*}[t]
\begin{center}
	\includegraphics[width=1.0\textwidth]{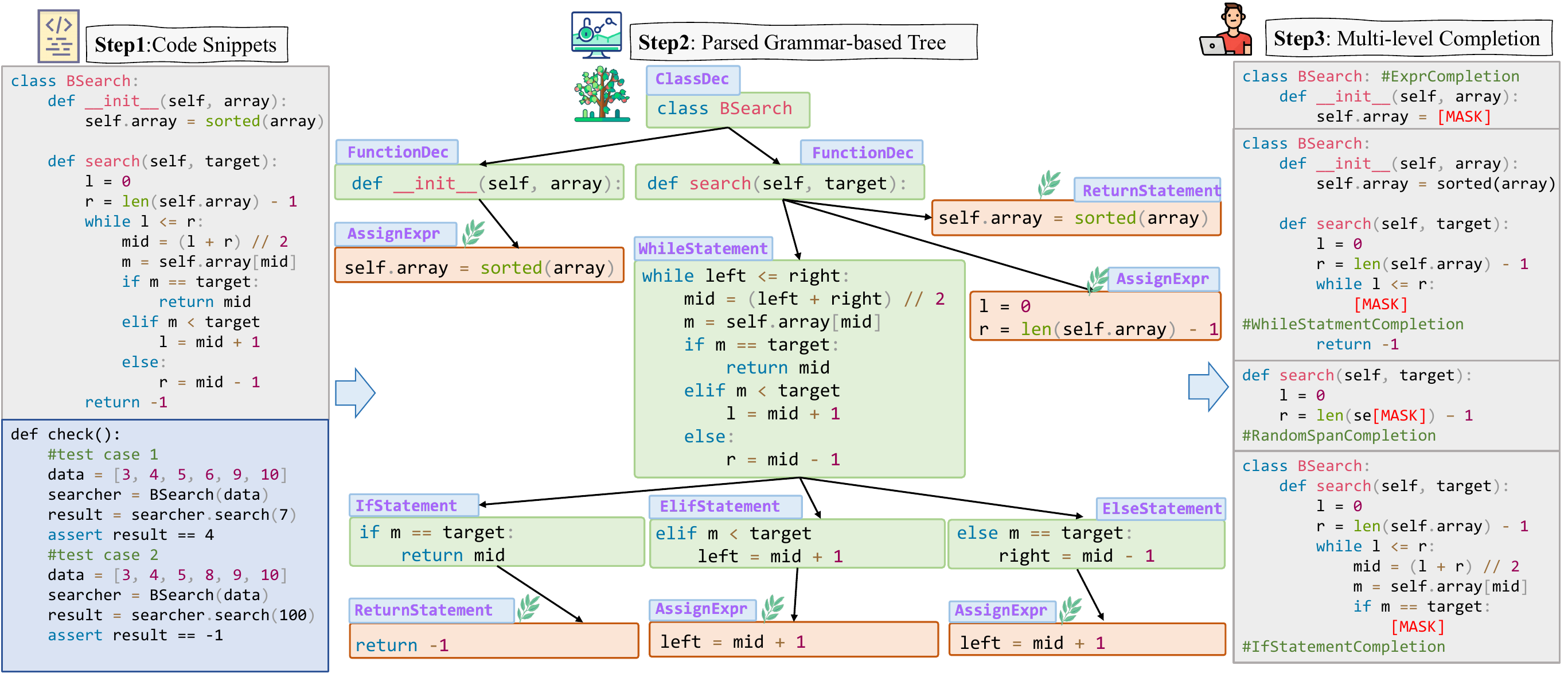}
	\caption{Multi-level Completion based on the parsed abstract syntax tree from the code snippet.}
	\label{model}
\end{center}
\end{figure*}

\section{\benchmark{} Construction}
\paragraph{Data Collection and Annotation}
The collected and refined repositories should follow the following guidelines:
(1) Search Github code repositories of the Python language that have been continuously updated. (2) Given the collected repositories, the annotator should collect or create the test cases for evaluation. (3) All collected repositories should pass the test cases in a limited time for fast evaluation (< 2 minutes). In Table \ref{repo_classification}, we collect diverse repositories for comprehensive code completion evaluation. We feed the prefix tokens and suffix tokens of the current file with the context tokens into the LLM to predict the middle code tokens.

\paragraph{Decontainmation.}
To avoid data leakage, we remove exact matches (20-gram word overlap) from CrossCodeEval~\cite{crosscodeeval} and the pre-training corpus stack V2~\citep{starcoder2}.

\begin{table*}[h!]
\centering
\resizebox{0.85\textwidth}{!}{
\begin{tabular}{lcccccc}
\toprule
& \multicolumn{3}{c}{Random Completion} & \multicolumn{3}{c}{Grammar-based Completion}  \\
& Span   & Single Line   & Multiple Line   & Expression     & Statement     & Function    \\ \midrule
$\lvert$Samples$\rvert$ & 42 & 34 & 38 & 407 & 266 & 377 \\
Context Tokens & 0/277.7K/27.7K & 333/276.7K/22.6K & 0/1484.1K/55.5K & 0/1484.4K/36.2K & 0/1484.6K/93.1K & 0/1484.5K/65.1K \\
Prefix Tokens  & 0/32.2K/1.4K & 0/38.3K/1.9K & 0/7.2K/978.0 & 0/35.8K/1.5K & 0/12.8K/786.0 & 0/38.3K/2.4K \\
Middle Tokens  & 3/22/7.0 & 4/48/15.0 & 4/156/40.0 & 2/150/13.0 & 2/74/9.0 & 7/123/33.0 \\
Suffix Tokens  & 0/7.9K/924.0 & 0/2.0K/562.0 & 0/5.8K/935.0 & 1/39.0K/1.3K & 1/40.1K/2.6K & 1/37.8K/2.0K \\ \midrule
  & $\lvert$Repositories$\rvert$ & $\lvert$Directories$\rvert$ & $\lvert$Stars$\rvert$ & $\lvert$Files$\rvert$ & $\lvert$Python Files$\rvert$ & $\lvert$Other Files$\rvert$ \\ 
Repository Overview  & 50 & 2/115/15 & 1/39K/2.6K & 15/790/113 & 4/411/38 & 10/379/75 \\
\bottomrule
\end{tabular}}
\caption{Data statistics of \benchmark{}.}
\vspace{-10pt}
\label{tab:benchmark_statistics}
\end{table*}

\paragraph{Data Statistics}
To create the benchmark \benchmark{}, we first construct the random span completion, random single-line completion, and random multi-line completion task by masking contiguous spans and lines of the chosen file of the whole repository. For the grammar-based completion, we first parse the code into an abstract syntax tree (AST) tree and randomly mask the node to match the input habits of programming developers habits.
Besides, we sort the context files using the relevance between the current masked file and truncate the tokens exceeding the maximum supported length of the code LLM.
The data statistic of \benchmark{} is listed in Table \ref{tab:benchmark_statistics}.

\section{\ourmethod{}}
\subsection{Problem Defintion}
\paragraph{In-file Completion}
Given the code $x^{L_{k}}$ of the current file of programming language $L_{k} (L_{k} \in L_{all}=\{L_{i=1}\}_{k=1}^{K})$, the LLM infills the middle code $x_{m}$ conditioned on the prefix code $x_{p}$ and the suffix code $x_{s}$ as follow:
\begin{MiddleEquation}
\begin{align}
    P(x_{m}^{L_{k}}) = P(x_{m}^{L_{k}}|x_{p}^{L_{k}},x_{s}^{L_{k}};\mathcal{M})
    \label{eval_code_in_file_completion}
\end{align}
\end{MiddleEquation}where $x^{L_{k}}_{p}$, $x^{L_{k}}_{s}$, and $x^{L_{k}}_{m}$ are concatenated as the complete code to be executed with the given test cases to verify the correctness of the generated code.

\paragraph{Repository-level Completion} Another more important completion scenario is the repository-level completion. Given the code snippets $z=\{z_{i=1}^{L_{k}}\}_{i=1}^{N}$ of $N$ other files, the LLMs try to fill the part code of the current file. Based on the prefix code of the current file $x_{p}^{L_{k}}$, suffix code of the current file $x_{s}^{L_{k}}$, and the code snippets $z=\{z_{i=1}^{L_{k}}\}_{i=1}^{N}$ in other files in the same repository, the LLMs aim at producing the middle code $x_{m}$ as:
\begin{MiddleEquation}
\begin{align}
    P(x_{m}^{L_{k}}) = P(x_{m}^{L_{k}}|x_{p}^{L_{k}},x_{s}^{L_{k}},\{z_{i=1}^{L_{k}}\}_{i=1}^{N};\mathcal{M})
    \label{eval_code_repo_completion}
\end{align}
\end{MiddleEquation}where the concatenation of $x^{L_{k}}_{p}$, $x^{L_{k}}_{s}$, and $x^{L_{k}}_{m}$ are used for repository-level execution for evaluation.

\subsection{Multi-level Grammar-based Completion}
Inspired by programming language syntax rules and user habits in practical scenarios, we leverage the \texttt{tree-sitter-languages}\footnote{\url{https://pypi.org/project/tree-sitter-languages/}} to parse the code snippets and extract the basic logic blocks as the middle code to infill. For example, the abstract syntax tree (AST) represents the structure of Python code in a tree format, where each node in the tree represents a construct occurring in the source code. The tree's hierarchical nature reflects the syntactic nesting of constructs in the code, and includes various elements such as expressions, statements, and functions. By traversing and manipulating the AST, we can randomly extract the nodes of multiple levels and use the code context of the same file to uncover the masked node.

\paragraph{Expression-level Completion}
At the expression level, we focus on completing sub-expressions within larger expressions or simple standalone expressions. This might involve filling in operand or operator gaps in binary operations or providing appropriate function arguments.

\paragraph{Statement-level Completion}
This level targets the completion of individual statements, such as variable assignments, control flow structures (if statements, for loops), and others. The goal is to maintain the logical flow and ensure syntactic correctness.

\paragraph{Function-level Completion}
At the function level, our approach involves completing entire function bodies or signature infillings. This includes parameter lists, return types, and the internal logic of the functions.

\section{Heuristic Completion Techniques}
To enhance the performance of our AST-based code infilling, we implement heuristic completion techniques to mimic the complementary habits of human users.

\paragraph{Random Line Completion}
We randomly select lines from the same file or similar files in the dataset to serve as candidates for completion. This process requires additional context-aware filtering to maintain relevance and accuracy.

\paragraph{Random Span Completion}
Instead of single lines, we randomly select code spans - sequences of lines that represent cohesive logical units. This approach suits larger blocks of code, needing a finer grasp of context and structure for effective completion.

\subsection{Hybrid Instruction Tuning}
Different from the base model trained with the FIM objective, we fine-tune the LLM with a mixture of the code completion data $(x_{p}^{L_{k}}, x_{m}^{L_{k}}, x_{s}^{L_{k}}) \in D_{x}=\{D_{x}^{L_{k}}\}_{k=1}^{K}$. The code completion training objective is described as:
\begin{MiddleEquation}
\begin{align}
    \mathcal{L}_{c} = -\frac{1}{K}\sum_{k=1}^{K}\mathbb{E}_{x_{p}^{L_{k}},x_{m}^{L_{k}},x_{s}^{k} \in D^{L_{k}}_{x}} [ P(x_{m}^{k}|x_{p},x_{s},z;\mathcal{M})
    \label{completion_sft}
\end{align}
\end{MiddleEquation}where the concatenation of $x^{L_{k}}_{p}$, $x^{L_{k}}_{s}$, and $x^{L_{k}}_{m}$ are used for repository-level execution for evaluation. $z=\{z_{i=1}^{L_{k}}\}_{i=1}^{N}$ are context code snippets of $N$ files in the same repository.

We also adopt the standard instruction data $(q^{L_{k}}, a^{L_{k}}) \in D_{q,a}=\{D_{q,a}^{L_{k}}\}_{k=1}^{K}$. The question-answer instruction tuning on $D_{q,a}$ is calculated by:
\begin{MiddleEquation}
\begin{align}
    \mathcal{L}_{qa} = -\frac{1}{K}\sum_{k=1}^{K}\mathbb{E}_{a^{L_{k}}, q^{L_{k}} \in D_{q,a}^{L_{k}}}\log P(q^{L_{k}}|a^{L_{k}};\mathcal{M})
    \label{qa_sft}
\end{align}
\end{MiddleEquation}where $(q^{L_{k}}, a^{L_{k}})$ are query and the corresponding response from the dataset $D_{x,y}$, including code generation, code summarization other code-related tasks. 

We unify the capability of the code completion and question-answer in a single instruction model. The training objective of the hybrid instruction tuning is described as:
\begin{MiddleEquation}
\begin{align}
    L_{all} = \mathcal{L}_{c} + \mathcal{L}_{qa}
    \label{all_sft}
\end{align}
\end{MiddleEquation}where $\mathcal{L}_{c}$ is the code completion objective and $\mathcal{L}_{qa}$ is the question-answering objective.

\section{Experiments}
\subsection{Code LLMs}
We evaluate 30+ models with sizes ranging from 0.5B to 30B+ parameters for open-source code large language models and closed-source general LLMs. For general models, we evaluate GPTs~\citep{gpt3,gpt4} (GPT-3.5-Turbo, GPT4-o) and Claude series~\cite{claude}. For code models, we test CodeLlama~\cite{code_llama}, StarCoder/StarCoder2~\citep{starcoder,starcoder2}, CodeGeeX~\cite{codegeex}, OpenCoder~\cite{opencoder}, Qwen-Coder~\citep{qwen25coder}, DeepSeekCoder~\citep{deepseek_coder}, CodeStral~\citep{codestral}, Yi-Coder\footnote{\url{https://huggingface.co/01-ai/Yi-Coder-9B}}, CodeGemma~\cite{codegemma}, and Granite-Coder~\cite{granite_coder}.


\subsection{Implementation Details}
We extract the repository-level code snippets from \texttt{the-stack-V2}\footnote{\url{https://huggingface.co/datasets/bigcode/the-stack-v2}} and filter the data with heuristic rules (e.g. GitHub stars and file length). We keep the mainstream programming language (Python, C-sharp, Cpp, Java, Javascript, Typescript, Php) and drop other long-tailed languages to obtain nearly 1.5M repositories. Finally, we obtain the instruction dataset \instruct{} contains nearly 3M completion samples. We fine-tune the open-source base foundation LLM Qwen2.5-Coder on nearly 3M instruction samples used in Qwen2.5-Coder \cite{qwen25coder} and code completion data (in-file and cross-file completion data). \ourmethod{} is fine-tuned on \texttt{Megatron-LM}\footnote{\url{https://github.com/NVIDIA/Megatron-LM}} with $64$ NVIDIA H100 GPUs. The learning rate first increases into $3 \times 10^{-4}$ with $100$ warmup steps and then adopts a cosine decay scheduler. We adopt the Adam optimizer~\cite{adam} with a global batch size of $2048$ samples, truncating sentences to $32$K tokens.

\subsection{Evaluation Metrics}
\paragraph{Edit Similarity} We compare the generated code and the ground-truth code using edit similarity (ES) to report string-based scores.

\paragraph{Pass@k} Similar to the in-file benchmark HumanEval/MBPP, we employ the Pass@k metric~\cite{codex} based on the executable results to get the reliability evaluation results. In this work, we report the greedy Pass@1 score of all LLMs with greedy inference for a fair comparison.

\subsection{Evaluation Benchmarks}

\paragraph{\benchmark{}} \benchmark{} is created with the repository-level unit tests to verify the correctness of the completion code, comprised of 1.2K samples from 50 active Python repositories. We separately report the ES score and Pass@1 in the table.

\paragraph{MultiPL-E} Since the mixture training of instruction samples and code completion samples, \ourmethod{} also supports answering the code-related queries. We adopt MultiPL-E \cite{multipl_e} for multilingual evaluation, including 8 popular programming languages.


\subsection{Main Results}

\begin{table*}[h!]
\centering
\resizebox{1.0\textwidth}{!}{
\begin{tabular}{lccccccccccccccc}
\toprule
\multirow{2}{*}{Models}             & \multirow{2}{*}{Params} & \multicolumn{6}{c}{Random Completion} & \multicolumn{6}{c}{Grammar-based Completion|} & \multicolumn{2}{c}{\multirow{2}{*}{Avg.}} \\
&    & \multicolumn{2}{c}{Span}   & \multicolumn{2}{c}{Single-line}   & \multicolumn{2}{c}{Multi-line}   & \multicolumn{2}{c}{Expression}     & \multicolumn{2}{c}{Statement}     & \multicolumn{2}{c}{Function}  & \multicolumn{2}{c}{}    \\ \midrule
&    & ES   & Pass@1   & ES   & Pass@1     & ES   & Pass@1   & ES   & Pass@1 & ES   & Pass@1 & ES   & Pass@1 & ES   & Pass@1 \\ \midrule
\rowcolor{cyan!15} Code-Llama & 7B &  3.7 & 11.9 & 6.8 & 35.3 & 17.2 & 26.3 & 5.8 & 28.5 & 5.9 & 23.3 & 17.3 & 15.6 & 9.9 & 22.7       \\
\rowcolor{cyan!15} Code-Llama  & 13B  & 3.4 & 19.0 & 6.5 & 35.3 & 16.7 & 26.3 & 5.7 & 29.5 & 6.3 & 25.6 & 17.4 & 17.5 & 9.9 & 24.4     \\ 
\rowcolor{cyan!15} Code-Llama  & 34B  & 4.5 & 9.5 & 6.5 & 32.4 & 16.9 & 18.4 & 6.7 & 28.7 & 6.5 & 22.9 & 17.8 & 16.7 & 10.5 & 22.6    \\
\rowcolor{cyan!15} Code-Llama  & 70B  & 3.9 & 16.7 & 6.8 & 38.2 & 17.7 & 26.3 & 5.8 & 28.7 & 6.1 & 25.6 & 17.6 & 19.9 & 10.0 & 24.9       \\ \midrule
\rowcolor{olive!15} Codestral   & 22B  &3.5 & 16.7 & 9.0 & 41.2 & 18.1 & 28.9 & 5.7 & 27.0 & 6.1 & 24.8 & 17.4 & 16.7 & 10.0 & 23.3        \\ \midrule
\rowcolor{orange!15} StarCoder   & 1B & 31.9 & 23.8 & 23.5 & 41.2 & 34.8 & 21.1 & 19.8 & 36.6 & 27.5 & 25.6 & 22.8 & 19.6 & 23.6 & 27.7 \\
\rowcolor{orange!15} StarCoder   & 3B & 16.6 & 11.9 & 11.7 & 41.2 & 24.6 & 26.3 & 12.2 & 28.3 & 18.0 & 26.7 & 19.6 & 15.1 & 16.5 & 23.4 \\
\rowcolor{orange!15} StarCoder   & 7B & 41.6 & 31.0 & 30.5 & 47.1 & 40.2 & 28.9 & 27.6 & 43.5 & 31.9 & 35.3 & 29.8 & 22.0 & 30.3 & 33.8 \\ \midrule
\rowcolor{lime!15} StarCoder2    & 3B & 2.9 & 14.3 & 5.8 & 38.2 & 15.1 & 21.1 & 4.8 & 26.3 & 5.3 & 21.4 & 15.0 & 15.1 & 8.5 & 21.3     \\
\rowcolor{lime!15} StarCoder2   & 7B  & 3.1 & 21.4 & 5.6 & 35.3 & 15.1 & 23.7 & 4.9 & 25.8 & 5.2 & 22.9 & 14.9 & 15.1 & 8.5 & 21.7      \\
\rowcolor{lime!15} StarCoder2  & 15B  & 2.7 & 19.0 & 5.6 & 41.2 & 14.8 & 23.7 & 4.9 & 27.5 & 5.2 & 23.7 & 15.1 & 15.6 & 8.5 & 22.8        \\ \midrule
\rowcolor{violet!15} DS-Coder    & 1.3B  & 35.0 & 21.4 & 25.7 & 47.1 & 32.1 & 28.9 & 27.2 & 32.9 & 14.9 & 27.1 & 27.6 & 14.1 & 24.9 & 25.3  \\ 
\rowcolor{violet!15} DS-Coder   & 6.7B & 40.2 & 28.6 & 41.0 & 50.0 & 47.5 & 39.5 & 45.7 & 37.3 & 36.1 & 33.8 & 45.9 & 15.1 & 43.3 & 29.5   \\ 
\rowcolor{violet!15} DS-Coder   & 33B  & 47.1 & 33.3 & 52.0 & 64.7 & 50.1 & 44.7 & 46.3 & 40.8 & 37.8 & 37.2 & 49.5 & 17.0 & 45.7 & 32.8 \\
\rowcolor{violet!15} DS-Coder-V2-Lite   & 2.4/16B  & 33.0 & 31.0 & 44.1 & 52.9 & 42.5 & 39.5 & 42.4 & 37.6 & 30.0 & 32.3 & 42.7 & 16.2 & 39.4 & 29.7 \\\midrule
\rowcolor{yellow!15} Granite-Coder    & 3B  & 35.0 & 21.4 & 25.7 & 47.1 & 32.1 & 28.9 & 27.2 & 32.9 & 14.9 & 27.1 & 27.6 & 14.1 & 24.9 & 25.3  \\ 
\rowcolor{yellow!15} Granite-Coder   &  8B & 0.0 & 19.0 & 0.0 & 58.8 & 2.6 & 28.9 & 5.2 & 36.6 & 0.0 & 26.3 & 0.0 & 21.5 & 1.9 & 29.1   \\ 
\rowcolor{yellow!15} Granite-Coder   &  20B  & 3.2 & 16.7 & 7.8 & 35.3 & 14.9 & 23.7 & 5.2 & 26.3 & 5.7 & 21.4 & 15.8 & 15.1 & 9.1 & 21.4 \\
\rowcolor{yellow!15} Granite-Coder   &  34B  & 3.1 & 16.7 & 8.0 & 35.3 & 15.2 & 26.3 & 5.3 & 26.3 & 6.2 & 24.1 & 15.4 & 15.6 & 9.1 & 22.3  \\ \midrule
\rowcolor{green!15}  CodeQwen1.5 & 7B  & 13.5 & 16.7 & 13.8 & 41.2 & 17.6 & 26.3 & 9.8 & 26.0 & 11.3 & 24.4 & 14.9 & 13.0 & 12.3 & 21.6\\
\rowcolor{green!15} Qwen2.5-Coder   & 0.5B & 11.3 & 16.7 & 10.4 & 47.1 & 13.0 & 26.3 & 10.7 & 26.0 & 14.6 & 24.1 & 16.2 & 14.1 & 13.5 & 22.0      \\
\rowcolor{green!15} Qwen2.5-Coder   & 1.5B & 3.5 & 14.3 & 3.2 & 29.4 & 7.9 & 15.8 & 4.0 & 21.9 & 3.5 & 16.9 & 9.1 & 11.7 & 5.6 & 17.2       \\
\rowcolor{green!15} Qwen2.5-Coder  & 3B & 13.8 & 19.0 & 14.9 & 44.1 & 12.5 & 21.1 & 13.9 & 28.0 & 11.2 & 23.7 & 18.5 & 13.5 & 14.8 & 22.3   \\ 
\rowcolor{green!15} Qwen2.5-Coder  & 7B & 7.8 & 16.7 & 10.2 & 35.3 & 12.4 & 23.7 & 5.3 & 24.3 & 8.1 & 21.1 & 11.0 & 12.5 & 8.3 & 19.8        \\ 
\rowcolor{green!15} Qwen2.5-Coder  & 14B & 7.0 & 16.7 & 12.7 & 35.3 & 12.1 & 18.4 & 11.4 & 27.5 & 15.1 & 27.8 & 18.5 & 13.5 & 14.4 & 22.6   \\ 
\rowcolor{green!15} Qwen2.5-Coder   & 32B & 3.9 & 16.7 & 32.5 & 47.1 & 20.3 & 23.7 & 21.2 & 29.5 & 26.1 & 33.5 & 33.0 & 15.4 & 25.8 & 25.7   \\ \midrule
\rowcolor{magenta!15} OpenCoder  & 1.5B & 1.7 & 11.9 & 3.4 & 38.2 & 5.4 & 23.7 & 3.2 & 26.3 & 3.2 & 27.4 & 6.6 & 15.4 & 4.3 & 22.8      \\ 
\rowcolor{magenta!15} OpenCoder  & 8B   & 2.7 & 14.3 & 4.4 & 32.4 & 10.8 & 21.1 & 4.0 & 29.5 & 3.7 & 24.1 & 7.8 & 16.7 & 5.3 & 23.4       \\ \midrule
\rowcolor{red!15} Yi-Coder  & 1.5B & 3.9 & 16.7 & 6.6 & 32.4 & 16.4 & 28.9 & 6.1 & 25.3 & 6.4 & 26.3 & 17.2 & 14.1 & 10.0 & 21.9      \\ 
\rowcolor{red!15} Yi-Coder  & 9B   & 3.4 & 16.7 & 6.8 & 29.4 & 17.7 & 26.3 & 5.8 & 28.0 & 6.3 & 26.3 & 17.6 & 17.8 & 10.1 & 23.9      \\ \midrule
\rowcolor{olive!15} CodeGemma  & 2B & 21.3 & 21.4 & 23.5 & 38.2 & 19.8 & 18.4 & 24.1 & 23.6 & 28.3 & 21.1 & 23.4 & 12.5 & 24.6 & 19.6    \\ 
\rowcolor{olive!15} CodeGemma  & 7B   & 12.4 & 19.0 & 14.3 & 35.3 & 28.2 & 26.3 & 15.4 & 29.2 & 18.7 & 36.5 & 25.8 & 18.3 & 19.8 & 27.1     \\ \midrule
\rowcolor{gray!15} \ourmethod{} & 7B  & 75.8 & 38.1 & 68.0 & 41.2 & 60.2 & 28.9 & 76.4 & 58.7 & 78.7 & 45.5 & 63.9 & 30.2 & 72.1 & 44.2 \\
\bottomrule
\end{tabular}}
\caption{Completion evaluation Results of base foundation model on \benchmark{}.}
\label{tab:execrepobench}
\end{table*}


\paragraph{\benchmark{}} Table \ref{tab:execrepobench} presents a comparative analysis of various code completion models, highlighting their performance across different metrics and parameter sizes. Code LLMs (e.g. CodeLlama and StarCoder) are evaluated across several completion tasks: random completion (span, single-line, multi-line), and grammar-based completion (expression, statement, function). Our proposed model \ourmethod{}, significantly outperforms competing models in all categories despite having only 7B parameters. Compared to the base foundation model Qwen2.5-Coder and DS-Coder, \ourmethod{} enhanced by the multi-level grammar-based fine-tuning achieves an impressive average score of 44.2, marking a substantial advancement in the field of code completion technologies. From the table, we can see that there exists a mismatch between the n-gram-based metric ES and execution-based metric pass@1. Granite-Coder-8B gets a good pass@1 score but a bad ES score, which emphasizes the importance of execution-based metric pass@k for correctly evaluating the code completion capability of code LLMs. ES metric has its own inherent flaws, where the score is calculated by the comparison between the generated code and ground-truth code. 

\paragraph{MultiPL-E} Table \ref{tab:coding_multiple} showcases the evaluation results in terms of Pass@1 performance (\%) across various models on the MultiPL-E benchmark, focusing on different programming languages. The comparison is categorically divided between proprietary models, like GPT-3.5 and GPT-4, and open-source models, which include DS-Coder, Yi-Coder, and Qwen2.5-Coder variants, among others. o1-preview, a proprietary model, leads with an average of 85.3\%, showcasing the difference in performance capability between proprietary and open-source models. The results highlight the effectiveness of our method, particularly in optimizing performance within the constraints of parameter size. Notably, our method, \ourmethod{}, with 7 billion parameters, outperforms other models in this parameter range across all listed programming languages, achieving an average Pass@1 performance of 76.4\%.

\section{Analysis}

\begin{figure*}[t]
\centering
    \subfigure[CrossCodeEval]{
    \includegraphics[width=0.4\textwidth]{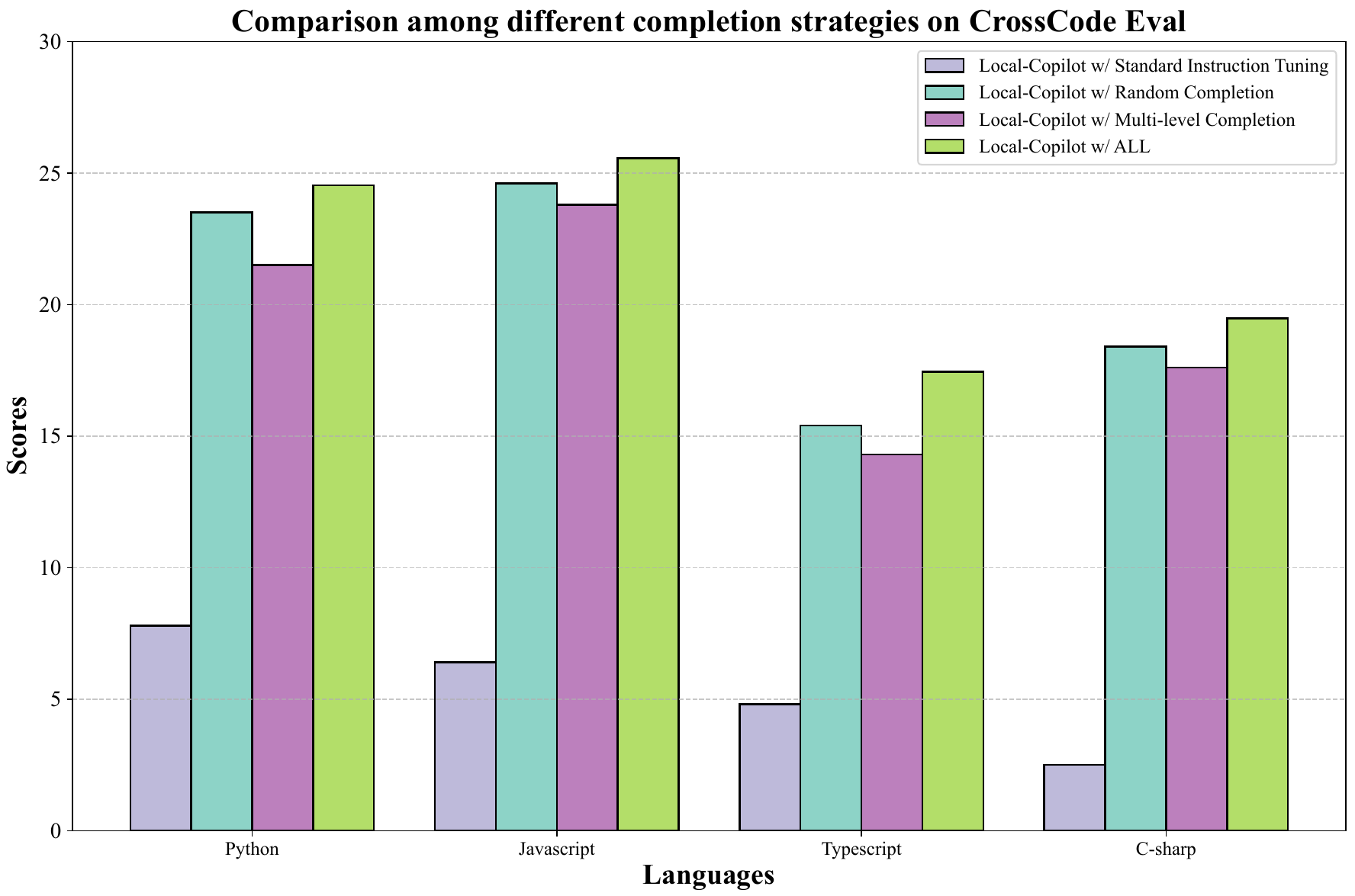}\quad
    \label{fig:ablation_study1}
    }
    \subfigure[MultiPl-E]{
    \includegraphics[width=0.4\textwidth]{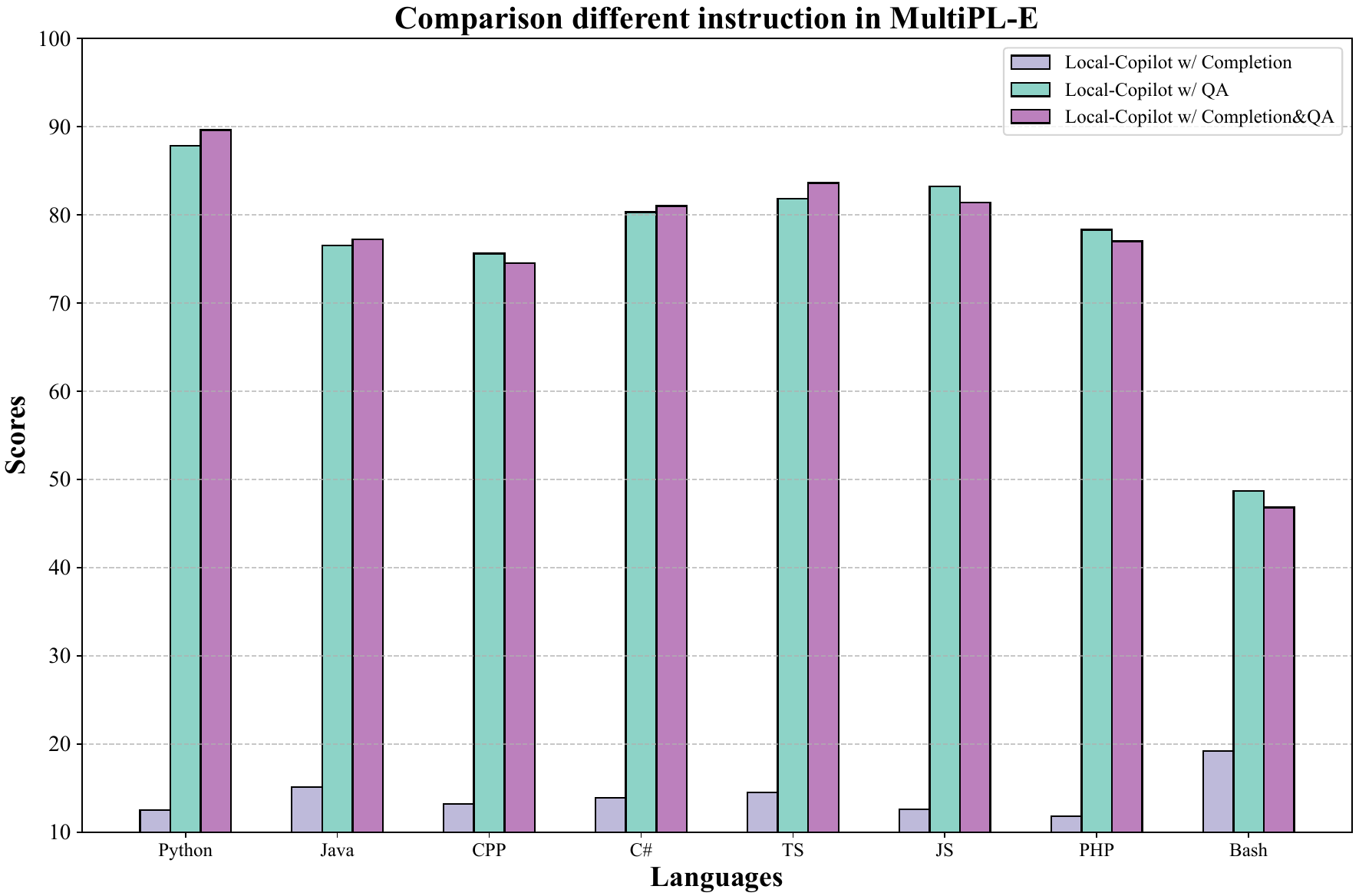}\quad
    \label{fig:ablation_study2}
    }
    \caption{Evaluation results based on standard QA pairs and code completion.}
    \label{fig:ablation_study}
    \vspace{-10pt}
\end{figure*}
\paragraph{Ablation Study}
Figure \ref{fig:ablation_study} emphasizes the essence of each component in our method by conducting the ablation study. CrossCodeEval~\cite{crosscodeeval} is developed using a variety of real-world, openly available repositories with permissive licenses, covering four widely used programming languages: Python, Java, TypeScript, and C-sharp. Figure \ref{fig:ablation_study1} shows the model results of the code completion task CrossCodeEval and Figure \ref{fig:ablation_study2} plots the results on the instruction-following code benchmark MultiPL-E. By unifying the code generation and completion in the same model, \ourmethod{} can support multiple scenarios.

\begin{table*}[h!]
 \centering
 \resizebox{0.85\textwidth}{!}{
 \begin{tabular}{lr|cccc|cccccccc|c}
 \toprule
 \textbf{Model} & \textbf{Size} & HE & HE+ & MBPP & MBPP+ & Python & Java & C++ & C\# & TS & JS & PHP & Bash & \textbf{Avg.} \\
 \midrule
 \multicolumn{15}{c}{\textbf{Closed-APIs}} \\ \midrule
 Claude-3.5-Sonnet-20240620 & \faLock{} & 89.0 & 81.1 & 87.6 & 72.0 & 89.6 & 86.1 & 82.6 & 85.4 & 84.3 & 84.5 & 80.7 & 48.1 & 80.2 \\
 Claude-3.5-Sonnet-20241022 & \faLock{} & 92.1 & 86.0 & 91.0 & 74.6 & 93.9 & 86.7 & 88.2 & \underline{87.3} & 88.1 & 91.3 & 82.6 & 52.5 & 83.8 \\
 GPT-4o-mini-2024-07-18 & \faLock{} & 87.8 & 84.8 & 86.0 & 72.2 & 87.2 & 75.9 & 77.6 & 79.7 & 79.2 & 81.4 & 75.2 & 43.7 & 75.0 \\
 GPT-4o-2024-08-06 & \faLock{} & 92.1 & 86.0 & 86.8 & 72.5 & 90.9 & 83.5 & 76.4 & 81.0 & 83.6 & 90.1 & 78.9 & 48.1 & 79.1 \\
 o1-mini & \faLock{} & \underline{97.6} & \underline{90.2} & \underline{93.9} & \underline{78.3} & 95.7 & \underline{90.5} & \underline{93.8} & 77.2 & \underline{91.2} & 92.5 & 84.5 & \underline{55.1} & 85.1 \\
 o1-preview & \faLock{} & 95.1 & 88.4 & 93.4 & 77.8 & \underline{96.3} & 88.0 & 91.9 & 84.2 & 90.6 & \underline{93.8} & \underline{90.1} & 47.5 & \underline{85.3} \\
 \midrule
 \multicolumn{15}{c}{\textbf{0.5B+ Models}} \\ \midrule
 Qwen2.5-Coder-0.5B-Instruct & 0.5B & \underline{61.6} & \underline{57.3} & 52.4 & \underline{43.7} & \underline{61.6} & \underline{57.3} & \underline{52.4} & \underline{43.7} & \underline{50.3} & \underline{50.3} & \underline{52.8} & \underline{27.8} & \underline{49.6} \\
 \midrule
 \multicolumn{11}{c}{\textbf{1B+ Models}} \\ \midrule
 DS-Coder-1.3B-Instruct & 1.3B & 65.9 & 60.4 & 65.3 & 54.8 & 65.2 & 51.9 & 45.3 & 55.1 & 59.7 & 52.2 & 45.3 & 12.7 & 48.4 \\
 Yi-Coder-1.5B-Chat & 1.5B & 69.5 & 64.0 & 65.9 & 57.7 & 67.7 & 51.9 & 49.1 & 57.6 & 57.9 & 59.6 & \underline{52.2} & 19.0 & 51.9 \\
 Qwen2.5-Coder-1.5B-Instruct & 1.5B & \underline{70.7} & \underline{66.5} & \underline{69.2} & \underline{59.4} & \underline{71.2} & \underline{55.7} & \underline{50.9} & \underline{64.6} & \underline{61.0} & \underline{62.1} & 59.0 & \underline{29.1} & \underline{56.7} \\
 \midrule
 \multicolumn{15}{c}{\textbf{3B+ Models}} \\ \midrule
 Qwen2.5-Coder-3B-Instruct & 3B & \underline{84.1} & \underline{80.5} & \underline{73.6} & \underline{62.4} & \underline{83.5} & \underline{74.7} & \underline{68.3} & \underline{78.5} & \underline{79.9} & \underline{75.2} & \underline{73.3} & \underline{43.0} & \underline{72.1} \\
 \midrule
 \multicolumn{15}{c}{\textbf{6B+ Models}} \\ \midrule
 CodeLlama-7B-Instruct & 7B & 40.9 & 33.5 & 54.0 & 44.4 & 34.8 & 30.4 & 31.1 & 21.6 & 32.7 & - & 28.6 & 10.1 & - \\
 DS-Coder-6.7B-Instruct & 6.7B & 74.4 & 71.3 & 74.9 & 65.6 & 78.6 & 68.4 & 63.4 & 72.8 & 67.2 & 72.7 & 68.9 & 36.7 & 66.1 \\
 CodeQwen1.5-7B-Chat & 7B & 83.5 & 78.7 & 77.7 & 67.2 & 84.1 & 73.4 & 74.5 & 77.8 & 71.7 & 75.2 & 70.8 & 39.2 & 70.8 \\
 Yi-Coder-9B-Chat & 9B & 82.3 & 74.4 & 82.0 & 69.0 & 85.4 & 76.0 & 67.7 & 76.6 & 72.3 & 78.9 & 72.1 & 45.6 & 71.8 \\
 DS-Coder-V2-Lite-Instruct & 2.4/16B & 81.1 & 75.6 & 82.8 & 70.4 & 81.1 & \underline{76.6} & \underline{75.8} & 76.6 & 80.5 & 77.6 & 74.5 & 43.0 & 73.2 \\
 Qwen2.5-Coder-7B-Instruct & 7B & \underline{88.4} & \underline{84.1} & \underline{83.5} & \underline{71.7} & \underline{87.8} & 76.5 & 75.6 & \underline{80.3} & \underline{81.8} & \underline{83.2} & \underline{78.3} & \underline{48.7} & \underline{76.5} \\
 OpenCoder-8B-Instruct & 8B & 83.5 & 78.7 & 79.1 & 69.0 & 83.5 & 72.2 & 61.5 & 75.9 & 78.0 & 79.5 & 73.3 & 44.3 & 71.0 \\
 \midrule
 \multicolumn{15}{c}{\textbf{13B+ Models}} \\ \midrule
 CodeLlama-13B-Instruct & 13B & 40.2 & 32.3 & 60.3 & 51.1 & 42.7 & 40.5 & 42.2 & 24.0 & 39.0 & - & 32.3 & 13.9 & - \\
 Starcoder2-15B-Instruct-v0.1 & 15B & 67.7 & 60.4 & 78.0 & 65.1 & 68.9 & 53.8 & 50.9 & 62.7 & 57.9 & 59.6 & 53.4 & 24.7 & 54.0 \\
 Qwen2.5-Coder-14B-Instruct & 14B & \underline{89.6} & \underline{87.2} & \underline{86.2} & \underline{72.8} & \underline{89.0} & \underline{79.7} & \underline{85.1} & \underline{84.2} & \underline{86.8} & \underline{84.5} & \underline{80.1} & \underline{47.5} & \underline{79.6} \\
 \midrule
 \multicolumn{15}{c}{\textbf{20B+ Models}} \\ \midrule
 CodeLlama-34B-Instruct & 34B & 48.2 & 40.2 & 61.1 & 50.5 & 41.5 & 43.7 & 45.3 & 31.0 & 40.3 & - & 36.6 & 19.6 & - \\
 CodeStral-22B-v0.1 & 22B & 81.1 & 73.2 & 78.2 & 62.2 & 81.1 & 63.3 & 65.2 & 43.7 & 68.6 & - & 68.9 & 42.4 & - \\
 DS-Coder-33B-Instruct & 33B & 81.1 & 75.0 & 80.4 & 70.1 & 79.3 & 73.4 & 68.9 & 74.1 & 67.9 & 73.9 & 72.7 & 43.0 & 69.2 \\
 CodeLlama-70B-Instruct & 70B & 72.0 & 65.9 & 77.8 & 64.6 & 67.8 & 58.2 & 53.4 & 36.7 & 39.0 & - & 58.4 & 29.7 & - \\
 DS-Coder-V2-Instruct & 21/236B & 85.4 & 82.3 & 89.4 & 75.1 & 90.2 & \underline{82.3} & \underline{84.8} & 82.3 & 83.0 & 84.5 & \underline{79.5} & \underline{52.5} & \underline{79.9} \\
 Qwen2.5-Coder-32B-Instruct & 32B & \underline{92.7} & 87.2 & 90.2 & 75.1 & \underline{92.7} & 80.4 & 79.5 & \underline{82.9} & \underline{86.8} & \underline{85.7} & 78.9 & 48.1 & 79.4 \\
 Qwen2.5-32B-Instruct & 32B & 87.8 & 82.9 & 86.8 & 70.9 & 88.4 & 80.4 & 81.0 & 74.5 & 83.5 & 82.4 & 78.3 & 46.8 & 76.9 \\
 Qwen2.5-72B-Instruct & 32B & 85.4 & 79.3 & \underline{90.5} & \underline{77.0} & 82.9 & 81.0 & 80.7 & 81.6 & 81.1 & 82.0 & 77.0 & 48.7 & 75.1 \\ 
 Qwen2.5-SynCoder & 32B & \underline{92.7} & \underline{87.8} & 86.2 & 74.7 & 92.1 & 80.4 & 80.7 & 81.6 & 83.0 & \underline{85.7} & 77.6 & 49.4 & 78.8 \\ \midrule
 \rowcolor{gray!15} \ourmethod{} & 7B & 87.2 & 81.1 & 81.7 & 68.5 & 89.6 & 77.2 & 74.5 & 81.0 & 83.6 & 81.4 & 77.0 & 46.8 & 76.4 \\
 \bottomrule
 \end{tabular}
 }
 \caption{The performance of different instruction LLMs on EvalPlus and MultiPL-E. ``HE'' denotes the HumanEval, ``HE+'' denotes the plus version with more test cases, and ``MBPP+'' denotes the plus version with more test cases.}
\label{tab:coding_multiple}
\end{table*}

\begin{figure}[h!]
\centering
\includegraphics[width=0.8\columnwidth]{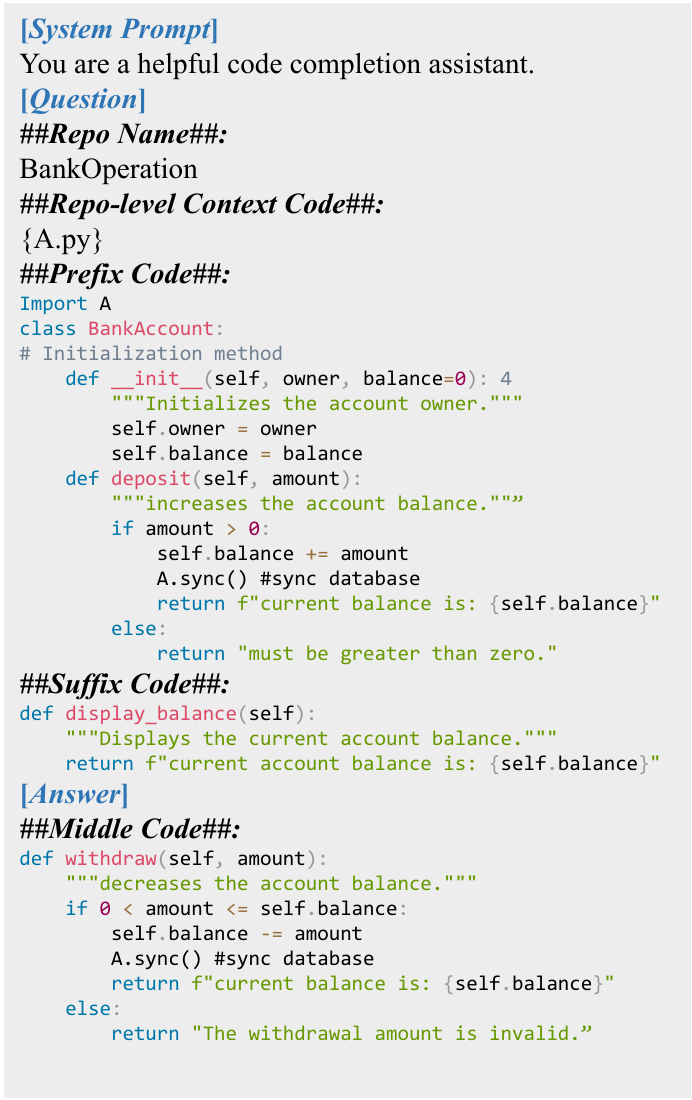}
\caption{Example of \ourmethod{}.}
\label{fig:example}
\caption{A completion example in instruction format of \ourmethod{}.}
\vspace{-15pt}
\end{figure}

\paragraph{Case Study}
Figure \ref{fig:example} showcases a part of a Python module named \texttt{BankOperation} which focuses on simulating basic bank account operations. The module, assumed to be spread across files, includes the BankAccount class definition housed within the given code snippet. Within this class, methods are defined for initializing an account (\texttt{\_\_init\_\_}), depositing money (deposit), and displaying the account balance (display\_balance). The core segment provided adds a withdraw method to this class, which allows for deducting a specified amount from the account's balance if the amount is positive and does not exceed the available balance. Each transaction (deposit and withdrawal) is followed up with a call to \texttt{A.sync()}, hinting at an operation to synchronize the current state of the account with a database, potentially managed by code within \texttt{A.py}. Error handling is incorporated within the deposit and withdrawal operations to ensure amounts are valid. The description wraps up this modular approach to implementing a banking system in Python, emphasizing object-oriented programming principles, error management, and database integration. This example shows that \ourmethod{} can successfully find the dependency from the context file.

\section{Related Work}
\paragraph{Code Large Language Models}
In software engineering, the advent of large language models (LLMs) tailored for code-centric tasks has proven to be transformative. Models~\cite{code_bert,codex,bloom,AlphaCode,santacoder,incoder,codet5,opencodeinterpreter,aixcoder,codegen2,magicoder,codegemma} like CodeLlama~\cite{code_llama}, DeepSeek-Coder \cite{deepseek_coder}, OpenCoder~\cite{opencoder} and Qwen2.5-Coder \cite{qwen25coder} — all trained on vast corpuses comprising billions of code snippets — have fundamentally augmented the development process. These Code LLMs are instrumental in automating repetitive software tasks, proposing code improvements, and facilitating the conversion of natural language into executable code. Notable among these are Starcoder \cite{starcoder, starcoder2}, CodeLlama \cite{code_llama}, and Code-Qwen \cite{Qwen}, each bringing unique contributions to the enhancement of coding assistance tools. With these advancements, Code LLMs showcase a promising trajectory for further revolutionizing how developers interact with code, promising ever-greater efficiency and intuitiveness in software creation. Inspired by the success of the grammar-based parsed tree in many fields, we adopt the abstract syntax tree to augment the code completion training.

\paragraph{Repo-level Code Evaluation}
In the domain of code evaluation, a rich tapestry of benchmarks \cite{codegeex,codereval,arcade_nl2code,humaneval_xl,xcodeeval,codeeditorbench,ds1000} has been woven to address the challenges of accurately assessing code quality, functionality, and efficiency, such as HumanEval/MBPP~\citep{codex,mbpp}, their upgraded version EvalPlus \cite{evalplus}, and the multilingual benchmark MultiPL-E \cite{multipl_e}, McEval~\cite{chai2024mceval}, and MdEval~\cite{mdeval}. BigCodeBench~\citep{bigcodebench}, fullstack~\cite{fullstack}, CodeFavor~\citep{codefavor}, CodeArena~\cite{codearena} and SAFIM~\cite{safim} separately evaluate code LLMs for more diverse scenarios and code preferences.
Studies have explored a variety of approaches, ranging from static analysis techniques (e.g. exact match (EM) and edit similarity (ES)), which examine code without executing it, to dynamic methods that involve code execution in controlled environments (e.g. Pass@k). The current benchmarks support code models to evaluate a series of different types of tasks, such as code understanding, code repair \cite{QuixBugs,debugbench,swe_bench}, code translation \cite{CodeTransOcean}, and multilingual scenarios \cite{odex,mbxp,codegeex,humaneval_xl,BabelCode}. An important task FIM \cite{incoder,fim,ding2024horizon} is to fill the middle code, given the prefix and suffix code, which provides substantial assistance for software development. 
Repo-level completion, such as RepoEval \cite{repoeval}, CrossCodeEval \cite{crosscodeeval,repoformer} and RepoBench \cite{repobench} only using exact match and edit similarity without code execution can not accurately reflect the model performance and Humaneval-Fim \cite{codegeex} focus in-file evaluation. 

\section{Conclusion}
In this work, we represent a significant leap forward in the realm of code completion, driven by the advancements in large language models (LLMs) tailored for coding tasks. By introducing an executable repository-level benchmark, \benchmark{}, alongside a novel, multi-level grammar-based instruction corpora, \instruct{}, the study not only tackles the limitations of existing benchmarks but also sets a new standard for evaluating code completion tools in real-world software development scenarios. The fine-tuning of base LLMs with 7B parameters, culminating in \ourmethod{}, demonstrates a remarkable improvement in code completion accuracy and efficiency across various programming languages, outperforming previous models. We pave the way for more sophisticated, accurate, and contextually aware code completion aids, promising to significantly enhance developer productivity, reduce error rates, and make software development more accessible and efficient for programmers worldwide.

\clearpage
\section{Limitations}
\label{sec:limitations}
We acknowledge the following limitations of this study: (1) The evaluation in repository-level multilingual scenarios are not fully explored. (2) The code completion model \ourmethod{} is mainly supervised fine-tuned on the 7B open-source base LLMs. In the future, we will try the (3) The fine-tuned model can be further improved using RLHF for better user experience, such as DPO.

\section*{Ethics Statement}
This research adheres to ethical guidelines for AI development. We aim to enhance the capabilities of large language models (LLMs) while acknowledging potential risks such as bias, misuse, and privacy concerns. To mitigate these, we advocate for transparency, rigorous bias testing, robust security measures, and human oversight in AI applications. Our goal is to contribute positively to the field and to encourage responsible AI development and deployment.
\bibliography{anthology,custom}

\clearpage

\end{document}